# A Modified Image Comparison Algorithm Using Histogram Features

**Anas M. Al-Oraiqat**
Taibah University, Department of Computer Sciences and Information
Kingdom of Saudi Arabia, P.O. Box 2898

**Natalya S. Kostyukova**
SHEE «Donetsk National Technical University»
Ukraine, P.O.Box 85300

*Abstract*—This article discuss the problem of color image content comparison. Particularly, methods of image content comparison are analyzed, restrictions of color histogram are described and a modified method of images content comparison is proposed. This method uses the color histograms and considers color locations. Testing and analyzing of based and modified algorithms are performed. The modified method shows 97% average precision for a collection containing about 700 images without loss of the advantages of based method, i.e. scale and rotation invariant.

## I. INTRODUCTION

In some tasks, the solution requires comparison of images. For example development of image and text recognizing systems and also special image searching systems. Image comparison is also required upon searching of odd copies in image collection and development of requests of databases that contain images. Thus, the task of comparison of color images should be solved since it is highly demanded in many fields.

There are many algorithms for comparison of images but all of them require a lot of time and considerable computing capacities and most of them are not efficient enough. However, the need for such algorithm increases continuously due to the expansion of multimedia technologies. The purpose of this article is to analyze a modified algorithm, offered by the authors, for comparing the content of color images in terms of the efficiency, spatial and temporal complexity. The proposed algorithm considers color characteristics of the image and spatial configuration of scene elements. It is resistant to most transformations such as scaling, turn and some changes of illumination.

## II. RELATED WORK

Today, digitalization and storage of large volume of visual materials is not a problem in technical sense. Actually, the main problem is to ensure an efficient and appropriate access to the relevant information in digital collection of images. Methods for searching images in databases can be divided into two types: searching by text description and searching by content. Searching by text attribute has some disadvantages; such as:
- The need of manually processing all images (it is almost impossible in viewing large number of images in the Internet);
- Ambiguity of text descriptions (it reduces accuracy and search completeness);
- The availability of images, which are hard to explain.

The other solution for image base retrieval problem is the Content Based Image Retrieval (CBIR). This term was used for the first time in 1992. Initially, it was the name of the method for images retrieval from database, based on some characteristics such as color, texture and element shape. Today CBIR is a combination of image retrieval technologies, based on content analysis.

Usually, CBIR-system functions in two stages: indexing and retrieval. Each image is described and entered into the database at indexing stage. It is important that not keywords or files are indexed, but also the main parameters of the image itself, analyzed with the help of special algorithms. As a rule, these parameters are color, texture and shape parameters. Obtained data are stored in index database. Then it is possible to retrieve images on the basis of definite values of such parameters. Parameters of one image are compared to similar data of other images, which are stored in index database, at retrieval stage.

General purpose of image retrieval is stated in [1, 19]. Various image feature extraction methods are proposed to solve the comparison problem. In particular, it is offered to use color features [2, 3], texture features [4, 5] and shape descriptors [6]. Various metrics may be used as a similarity or difference measure. The comparison of popular distance measures is represented by [7].

Normally, characteristic of image content is a vector. Two groups of methods are used for image comparison. The first group includes methods that use global features. General features give general description: the average brightness, the average value for separate channels etc. The second group is represented by methods that use local features. Local features describe image part, for instance, the average brightness in upper left quarter and the average value for red channel in the vicinity of image center. However, color, texture, and shape should be considered either for global or local images upon comparison of images. As a rule, colorimetric quantities are used in algorithms to have more efficiency. Texture and shape feature extraction process is more time-consuming and less clear. For instance, it is quite easy to determine the color of one pixel since it is always set but it is impossible to determine the texture with one pixel and it is necessary to consider neighborhood for each pixel.

If global features are used, perceptual cash method and color histogram methods are the most common.

Perceptual cash algorithms [8-10] describe function class for generation of individual (non-unique) print (cash) and these prints can be compared to each other. Perceptual cash is a development of cryptographic cash function concept. Perceptual cash calculation algorithms have the following basic features: it is possible to change image size, aspect ratio and even color features (brightness, contrast etc.), but they still will coincide by cashes [11].

Methods, based on use of color histograms, are also quite popular. There are various methods for plotting and comparing color histograms [12-15] that differ from each other by initial color space (RGB, CMY, HSV, grayscale), histogram size and determination of the distance between histograms. Correlation method [16], chi-quart method [16], histogram intersection calculation and Kullback-Leibler distance [16] are used for comparison of histograms. Also the method using fuzzy computation for calculation and comparison of color histograms showed good results [21]. Histogram intersection and chi-quart method are most commonly used in practice. The disadvantage of color histogram method is its sensitiveness to color localization, which may lead to false results.

III. THE PROPOSED ALGORITHM

Color location algorithm, proposed by authors in this paper, is a modification of the color histogram method. Let's consider algorithm steps in details.

A. Step 1. Image preprocessing

The image is preprocessed before execution of the algorithm. This preprocessing removes high-frequency components of the image by reduction.

The first step of the algorithm is to plot color histogram in HSV space [16]. The use of HSV color space ensures the resistance to image shading. Color space quantization is performed before calculation of color histogram elements. The authors propose to perform uneven quantization that takes into account the peculiarities of human perception of color. When quantizing, more intervals are allocated for the H component reflecting the hue and fewer gaps for the S and V components. Quantization for the color tone component is performed for such discrete levels: 16, 26, 40, 70, 85, 145, 160, 220, 262, 278 and 335 nm.

Density and brightness component quantization is also non-uniform. In general, the number of color is equal to 47 after quantization. Color histogram elements are calculated exactly for this quantity. It should be noted that histogram element normalizing procedure is compulsory.

Authors used the features of human color perception for choice of not uniform quantization of HSV color space. Also different sensitivities of human eyes to different values of tone were using. Quantization of saturation and hue is not uniform due to hues close to black are indistinguishable for human [16]. We consider the color black when V value is less than 0.2, regardless of H and S values. Also we consider the color grey when S value is less than 0,2, regardless of H and V values. Also we select the white color area which is limited by V=0,85 and H=0,1. Remaining color subspace is perceived by human as colored. The subspace is divided to four subregions with borders S = 0,65 and V = 0,7. So, the total number of colors after quantization is 47.

B. Step 2. Calculation of image characteristics

The second step of this algorithm is to calculate the characteristics, including color localization (color location). These characteristics have to be invariant with respect to transformations: image turn, displacement and zooming. Histograms should be preliminary rejected in descending order, based on the share of pixels of definite color on the image. Then "mass center" is calculated for certain number of first columns of histogram. Mass center is determined by calculating arithmetical mean values of point's position that have certain color. Figure 1 shows centers of five colors, which share is the largest one, and three-color set, for which colors from five color locations of the other image will be matched upon comparison. The creation of modified color histogram, which contains not only information about number of pixels of each color but also information about spatial position of each main color, is completed at this stage.

It is proposed to use formula, recommended by CIE International Committee (French: Commission Internationale de l'Eclairage), using coordinates ($L_1$, $a_1$, $b_1$) and ($L_2$, $a_2$, $b_2$) in color space Lab [16]:

$$dE = \sqrt{(L_2 - L_1)^2 + (a_2 - a_1)^2 + (b_2 - b_1)^2} \quad (1)$$

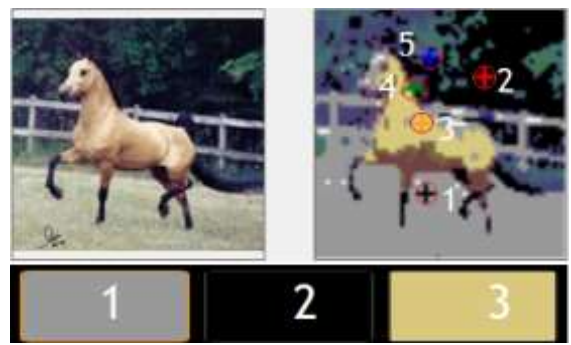

Figure 1. Image color locations and the set of three basic colors.

The result, obtained during basic comparison of color histograms, is also considered in the following way when comparing modified color histograms:

$$d(H_1, H_2) = k \cdot d_{hist}(H_1, H_2) + (1-k) \cdot d_{colorLoc}(H_1, H_2) \quad (2)$$

Where $k$ is the weighting factor ($0 \leq k \leq 1$);
$d_{hist}$ is a result of modified histogram comparison by basic method;
$d_{colorLoc}$ is a result of color location comparison.

The calculation of the difference between color locations of different images is the main stage of modified algorithm that comprises the content of color images. Then image color locations, taken from database, are sorted against image-request color locations.

Characteristic value of the distance is calculated for each color location by the following formula:

$$d_{colLoc0}(L_1, L_2) = 2 \cdot \frac{d_{point}(L_1, L_2)}{\max Dist} + d_{color}(L_1, L_2), \quad (3)$$

Where $L_1$ and $L_2$ are hypothetical and actual color locations, respectively;
$d_{point}$ and $d_{color}$ are functions, used to calculate the distance between points and colors;
$maxDist$ is the maximum distance between points.

Generally, difference between images is calculated as a sum of the distances for each color location. Sometimes the image from the base contains less number of colors than that needed for comparison after quantization. In this case the value of 10000 is assigned to corresponding distance. Thus, corresponding image will be transferred to the end of the list of similar images upon result's sorting. It is necessary to convert the color from HSV space to Lab space before calculation of the distance. Once such conversion is completed, the comparison will be performed by the formula (1).

Now, modified algorithm can be represented in the form of the following sequence of actions for image-request:
- Calculation of color histogram;
- Sorting histogram in decreasing order on the basis of pixel number;
- Calculation of color locations for the first N colors in histogram;
- Sorting of the list of image color locations from the base in accordance with the list of image-request color locations;
- Calculation of affine transformation coefficients on the basis of two first elements from the list of color locations of both images;
- Calculation of the distance between hypothetical and actual coordinates of the following color location in the list (the point, calculated for coordinate of image-query color location after transformation, is used as hypothetical coordinate and coordinates of corresponding color location from image base are used as actual coordinates);
- Calculation of color difference of corresponding color locations in the list.

Common sequence of the operation of software system, used for comparison of color images on the basis of algorithm offered by authors is shown on Figure 2.

## IV. EXPERIMENTAL RESULTS

The proposed method is compared with methods proposed by Liang's, Swain's and Ballard's, Tico's and with fuzzy linking histogram creation method. The estimation results of these methods are taken from [21].

Image database used by authors in [21] is not available now. So, we have created own test database. This database consists of about 700 various images. The images are downloaded from different web-sites or taken by camera. The database contains some groups of visually similar images. This fact allows us estimate characteristics of image retrieval. For example, figure 3 shows one of groups of similar images – photos of pink flower, centered on green natural background.

We use two characteristics of retrieval effectiveness: precision and recall [20]. Precision is the proportion of the retrieved images that are relevant to the query. Recall shows the proportion of relevant images in the entire database that are retrieved in the query. Let a, b and c are given as follows:
a – number of retrieved and relevant images;
b – number of retrieved and not relevant images;
c – number of not retrieved and relevant images.

Recall and precision are defined by the following formulas [20]:

$$\text{Recall} = \frac{a}{a+c}, \quad (4)$$

$$\text{Precision} = \frac{a}{a+b}. \quad (5)$$

Figure 4 shows a plot created by some experiments. As we can see, Recall values do not exceed Precision values in all experiments. It confirms that the proposed method is effective [21].

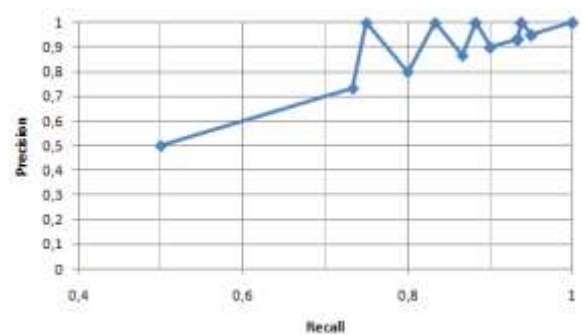

Figure 4. Experimental results of the proposed method

Because authors use own image databases, for comparison with results in [21] average Precision and Recall were calculated by data from [21] and by author's results. These values are shown in Figure 5. We can see that proposed method is a little bit worse than FuzzyHistogram method, but it is not worse than other traditional methods.

The search results are below. The query image is located first, in the upper left corner, the remaining images are located in order of similarity descending from left to right and top to bottom.

First search result is shown in Figure 6. The target of search was an image with a gray object (bird) in center and with plant grayish-green background.

As we can see, only one image (the first in the fifth row) does not look like a query image from the human point of view, however for the algorithm it contains the same colors in the same locations, therefore it is included in the search results. The last image differs from the query image by the set of colors and their locations, so it is placed in the end of the search results. For this example, Precision = 15/15 = 1. The total number of photos of a gray bird on a plant background in the collection is 16, that is, in this case Recall = 15/16 = 0.94.

Figure 7 also shows the retrieval results. The query image is a photo of yellow flower, located in the center of the image, on a plant background. As we can see, only one photo (the last one) does not look like a query image. In this case, Precision = 14/15 = 0.93. The total number of photos of the yellow flower in the collection is 17, that is in this case Recall = 14/17 = 0.82.

V. CONCLUSIONS

This article outlines and analyses corresponding methods for color image comparison content on the basis of local and global features. It also describes limitations of color histogram methods. The selection of HSV color space model and HSV color space non-uniform quantization diagram are justified in this study. This article also proposes a new algorithm for the calculation of spatial position of colors on the image, i.e. color location method, describes this algorithm and main steps for execution of offered modified algorithm. The efficiency of basic and modified methods have been tested and analyzed. The proposed method shows quite good results and can be used practically.

ACKNOWLEDGMENT

The authors would like to thank Taibah University and Donetsk National Technical University for supporting this research.

REFERENCE

1) E. A. Bashkov & Kostiukiva, "The evaluation of the image retrieval efficiency with use of 2d-color histograms," Management and Informatics Problem, No. 6, pp. 84-89, 2006.
2) N. S. Vasileva & B. A. Novikov, "The construction of correspondences between low-level characteristics & semantics of static images, "Electronic Libraries: Perspective Methods and Technologies - Electronic Collections," Works of The 7th All-Russian Scientific Conference, Yaroslavl, 2005.
3) M. J. Swain & D. H. Ballard, "Color Indexing," International Journal of Computer Vision, Vol. 7(1), pp. 11-32, 1991.
4) H. Tamura, S. Mori & T. Yamawaki, "Textural features corresponding to visual perception," IEEE Systems, Man, and Cybernetics Society, Vol. 8(6), pp. 460-473, 1978.
5) B. S. Manjunath & W. Y. Ma, "Texture features for browsing and retrieval of image data," IEEE Transactions of Pattern Analysis and Machine Intelligence, Vol. 18(8), pp. 837-842, 1996.
6) D. Zhang & G. Lu, "Content-Based Shape Retrieval Using Different Shape Descriptors," A Comparative Study, In IEEE International Conference on Multimedia and Expo, pp. 289-293, 2001.
7) Y. Rubner & C. Tomasi, "A Metric for Distributions with Applications to Image Databases," In Proceeding of the Sixth International Conference on Computer Vision, IEEE Computer Society, p. 59, 1998.
8) I. V.Rudakov & I. M. Vasiutovich, "The Study of Perceptive Cash-Functions of the Images," Science and Education, Journal of Moscow State Technical University named after Bauman, No.8, 2015.
9) L. E. Chalaya & P. Yu. Popadenko, "Search for Incomplete Copies in Digital Image Analysis System," Bulleting of Kremenchug National University named after Mykhailo Ostrogradskiy, No. 5(88), pp. 42- 47, 2014.
10) G. D. Ognevoy, "Methods and Algorithms for Image Retrieval," The IInd International Scientific and Technical Internet, Conference, 2014, https://rep.bntu.by/handle/data/12606, (Accessed on 21 Aug, 2017).
11) "Looks alike," Principles of Perceptive cash, http://habrahabr.ru/post/120562/, (Accessed on 15 Sep., 2017).
12) E. Ardizzone, M. La Cascia & D. Molinell, "Motion and Color Based Video Indexing and Retrieval," Proceedings of the 13th International Conference on, Pattern Recognition, 1996.
13) E. Ardizzone, M. La Cascia, Vito di Gesu & C. Valenti, "Content Based Indexing of Image and Video Databases by Clobal and Shape Features," 1996.
14) N. S. Baigarova, Yu. A. Bukhshta & A. A. Gornyi, "Visual Data Indexation and Retrieval Methods," Preprint of Application Mathematics Institute named after M. V. Kledysh, Russian Academy of Science,

No.7, 2000.
15) J. R. Smith & Shih-Fu Chang "Tools and Techniques for Color Image Retrieval," Columbia University Department of Electrical Engineering and Center for Telecommunications Research, 1996, http://www.ee.columbia.edu/ln/dvmm/ publications/96/smith96b.pdf, (Accessed on 10 Oct., 2017).
16) Shapiro L. & J. Stockman, "Computer Vision," Binom, Laboratory of Knowledge, pp. 763, 2009.
17) Microsoft Visual Studio 2013, https://www.visualstudio.com/en-us/downloads/download-visual-studio-vs.aspx, (Accessed on 20 Oct., 2017).
18) Microsoft.NET Framework 4.5, https://www.microsoft.com/ru-ru/download/details.aspx?id=30653, (Accessed on 05 Nov., 2017).
19) E. A. Bashkov, O. L. Vovk & N. S. Kostukova, "Content-Based Image Retrieval in Graphical Databases: monog," Public Higher Education Institution, p.120, Donetsk, 2014.
20) J. R. Smith, "Integrated Spatial and Feature Image Systems: Retrieval," Analysis and Compression: Submitted in partial fulfillment of the requirements for the degree of Doctor of Philosophy in the Graduate School of Arts and Sciences, Columbia University, 1997.
21) K. Konstantinidis, A. Gasteratos & I. Andreadis, "Image retrieval based on fuzzy color histogram processing," Optics Communications 248, pp. 375-386, 2005.

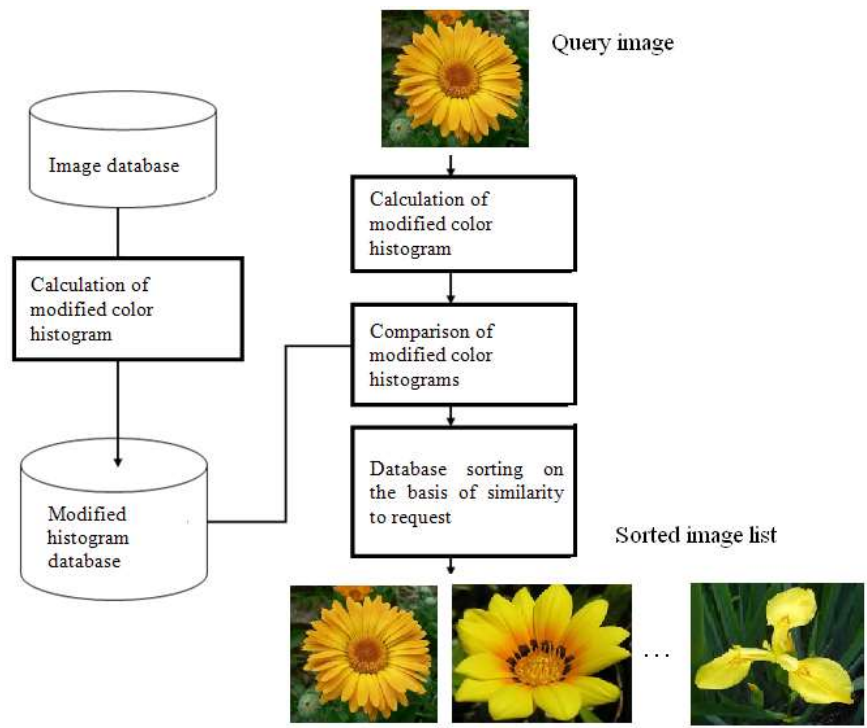

Figure 2. General diagrams for comparison of color image content algorithm.

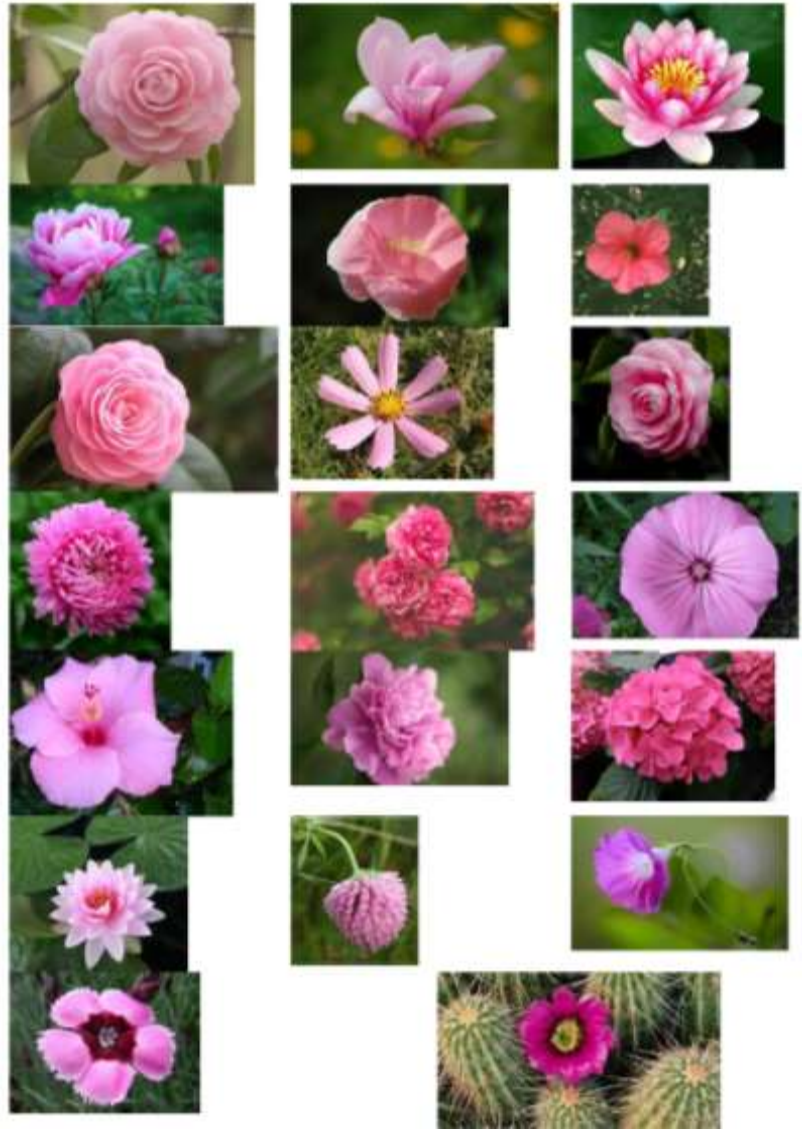

Figure 3. Group of similar images from test database.

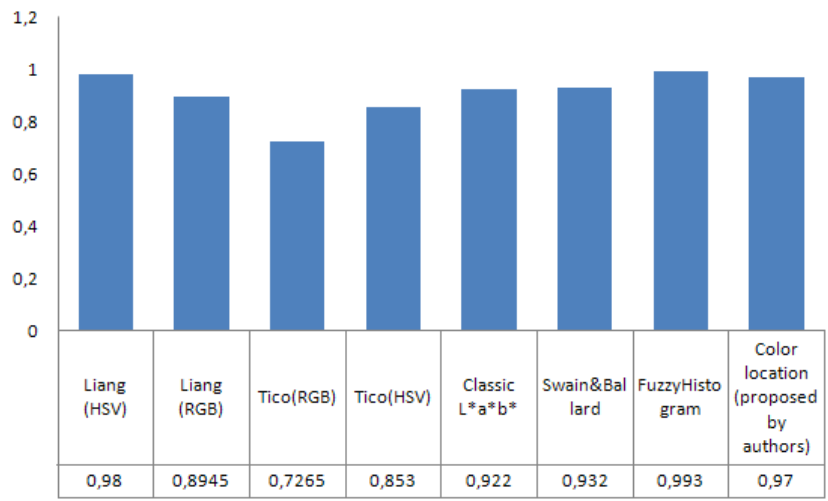

| | Liang (HSV) | Liang (RGB) | Tico(RGB) | Tico(HSV) | Classic L*a*b* | Swain&Ballard | FuzzyHistogram | Color location (proposed by authors) |
|---|---|---|---|---|---|---|---|---|
| | 0,98 | 0,8945 | 0,7265 | 0,853 | 0,922 | 0,932 | 0,993 | 0,97 |

Figure 5. Comparison of average Precision for different retrieval methods.

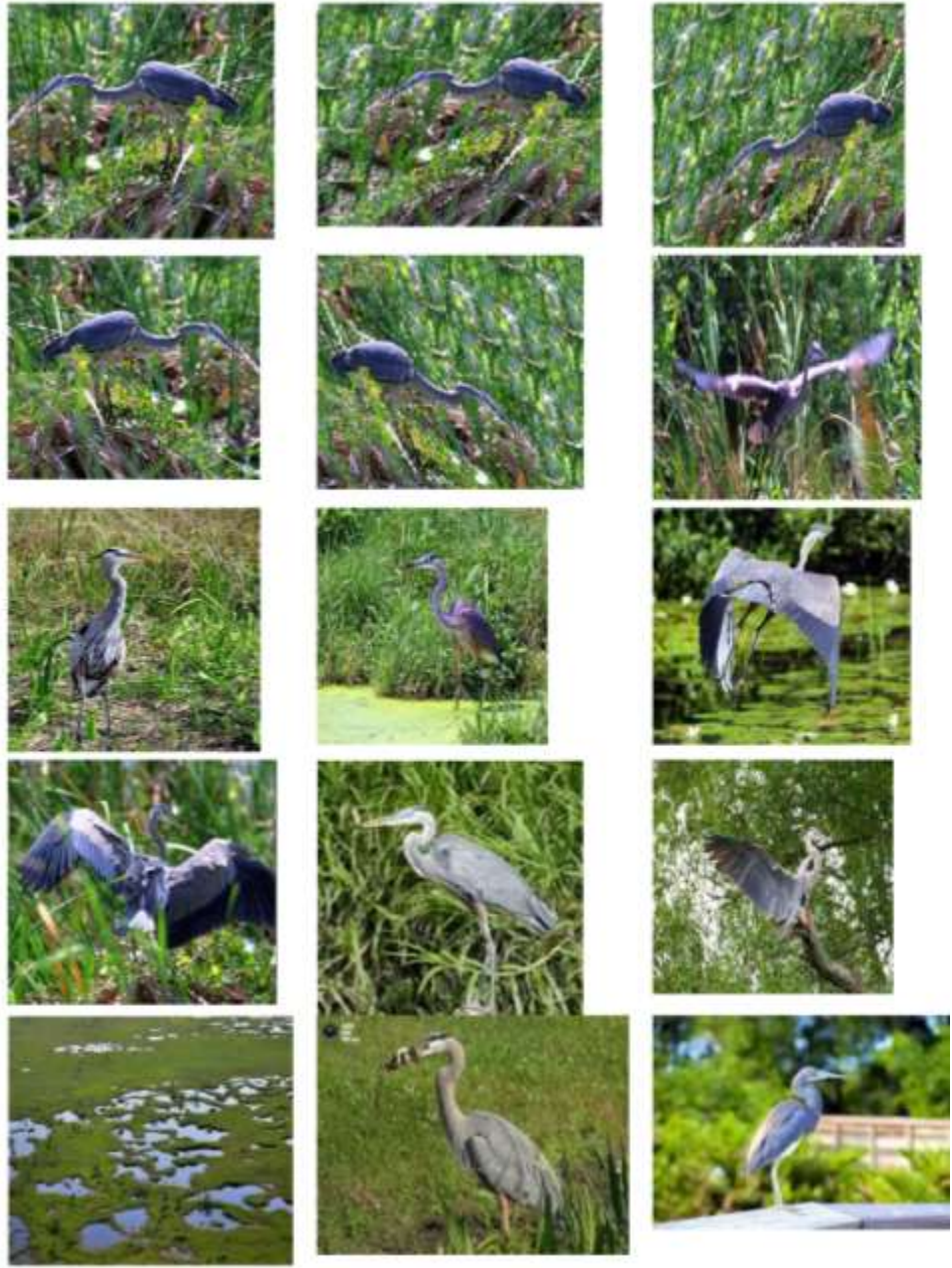

Figure 6. Image retrieval results (test 1).

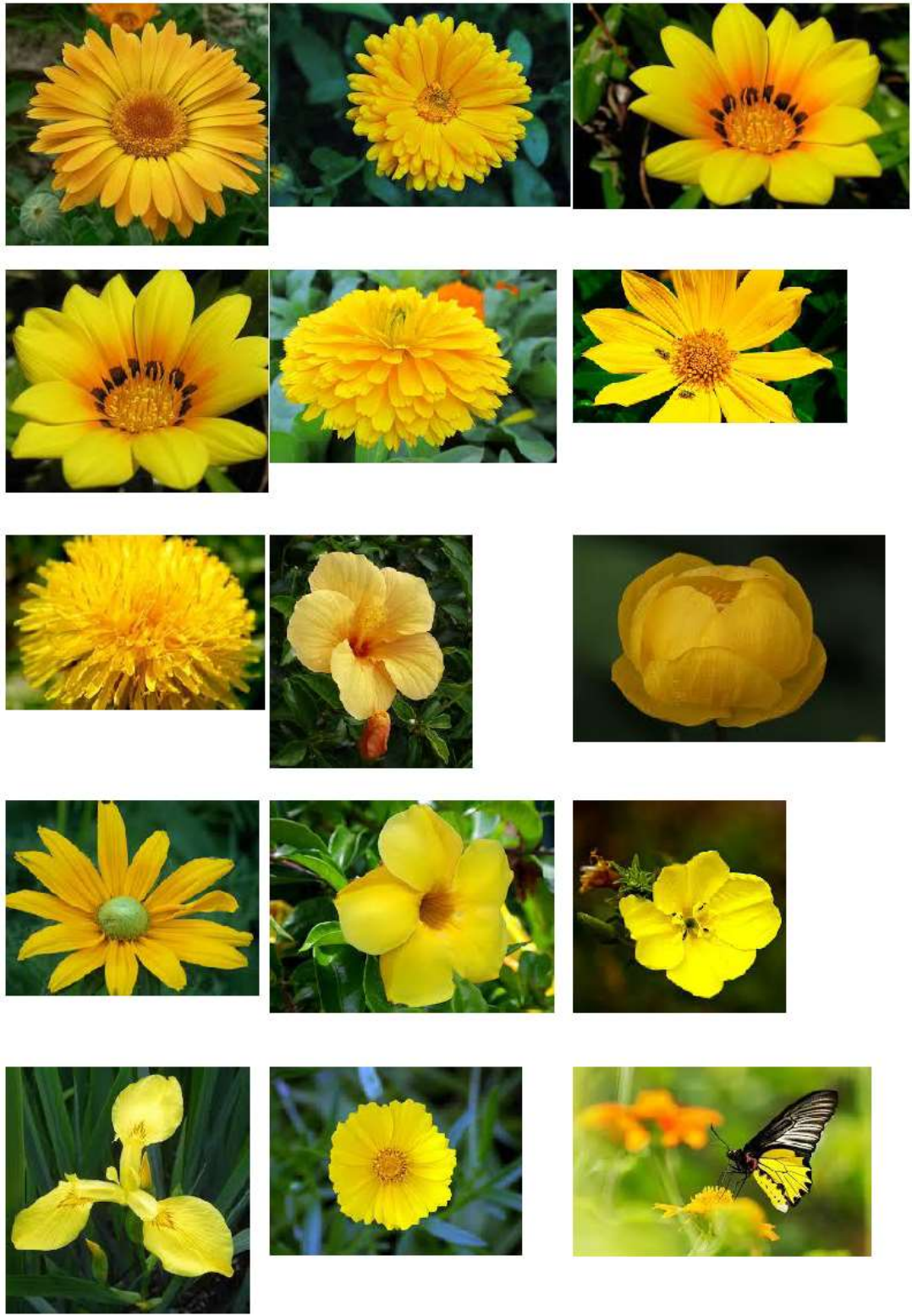

Figure 7. Image retrieval results (test 2).